# Automatic Detection of Diabetes Diagnosis using Feature Weighted Support Vector Machines based on Mutual Information and Modified Cuckoo Search

*Davar Giveki, Hamid Salimi, GholamReza Bahmanyar, Younes Khademian*

Department of Computer Sciences, University of Saarland, Saarbrucken, Germany
Email: s9dagive@stud.uni-saarland.de

Department of Computer Sciences, University of Tehran, Tehran, Iran
Email: salimi.hamid86@gmail.com

Department of Computer Sciences, University of Saarland, Saarbrucken, Germany
Email: s9ghbahm@stud.uni-saarland.de

Department of Electrical and Biomedical Engineering, Iran University of Science &Technology, Tehran, Iran
Email: khademian2001@gmail.com

**Abstract:** Diabetes is a major health problem in both developing and developed countries and its incidence is rising dramatically. In this study, we investigate a novel automatic approach to diagnose Diabetes disease based on Feature Weighted Support Vector Machines (FW-SVMs) and Modified Cuckoo Search (MCS). The proposed model consists of three stages: Firstly, PCA is applied to select an optimal subset of features out of set of all the features. Secondly, Mutual Information is employed to construct the FWSVM by weighting different features based on their degree of importance. Finally, since parameter selection plays a vital role in classification accuracy of SVMs, MCS is applied to select the best parameter values. The proposed MI-MCS-FWSVM method obtains 93.58% accuracy on UCI dataset. The experimental results demonstrate that our method outperforms the previous methods by not only giving more accurate results but also significantly speeding up the classification procedure.

**1 Introduction**
Diabetes disease diagnosis via proper interpretation of the Diabetes data is an important classification problem. Diabetes occurs when a body is unable to produce or respond properly to insulin which is needed to regulate glucose. Diabetes not only is a contributing factor to heart disease, but also increases the risks of developing Kidney disease, Blindness, Nerve damage, and blood vessel damage. Statistics show that more than 80 percent of people with Diabetes die from some form of heart or blood vessel diseases. Currently there is no cure for Diabetes; however, it can be controlled by injecting insulin, changing eating habits, and doing physical exercises (Polat and Güneş, 2007). Diabetes disease diagnosis via proper interpretation of the Diabetes data is an important classification problem.
Support Vector Machines (SVMs), developed by Vepnik (Vapnik, 1995), has been studied increasingly in recent years. It was applied to the problem of diagnosis of Diabetes diseases in several works (Çalisir and Dogantekin, 2011; Polat et al., 2008; Übeyli, 2007) due to outstanding characteristics and excellent generalization performance. Moreover, SVMs can handle nonlinear classification tasks efficiently when samples are mapped into higher dimensional feature space by a nonlinear kernel function. In order to determine an efficient classification technique for diagnosis problem, authors in (Übeyli et al.) compared SVM with a number of quantitative methods including: Combined Neural Networks (CNNs), Mixture of Experts (MEs), Multilayer Perceptrons (MLPs), Probabilistic Neural Networks (PNNs) and Recurrent Neural Network (RNN). They demonstrated that SVM is significantly more robust than the other methods. In addition, they showed that Radial Basis Functions (RBFs) are more efficient than Polynomial and Linear Kernel Functions in Diabetes diagnosis case. Therefore, we decided to use SVM classifier with RBF kernel function in our study.

Feature selection strategies often are applied to explore the effect of irrelevant features on the performance of classifier systems (Acır et al., 2006; Lin et al., 2008; Valentini et al., 2004; Zhang et al., 2005). In this phase, an optimal subset of features which are necessary and sufficient for solving a problem is selected. Feature selection improves the accuracy of algorithms by reducing the dimensionality and removing irrelevant features (Karabatak and Ince, 2009; Mehmet Fatih, 2009). In our task, we employed Principal Component Analysis (PCA) as feature selection strategy to make the classifier more effective. In recent years, PCA has been used in various fields such as: image recognition, signal processing, face recognition and image compression, etc. This method is a general technique for finding patterns in data. Here, PCA reduces the dimension of Diabetes disease dataset form 8 to 4 features. Consequently, we will only use 4 features in training and testing our proposed method.

SVMs are applicable to a wide variety of classification and recognition tasks with different datasets. Although in each dataset, there only some of the features are more relevant to a given classification task, all the existing SVMs assume that all the features have equal relevancy weight. To increase the effect of relevant features, it is desirable to propose an SVM method which weight features based on their degree of importance in a given classification task (Xing et al., 2009). Xing et al. (Xing et al., 2009) showed that Mutual Information (MI) can be used to weight features by their relevancy.

In our work, we use SVM with weighted features as the classification technique, where the features are weighted based on their relevancy by computing MI.

Choosing two parameters $C$ and $\gamma$ is very important in using SVMs. Parameter $C$ represents the cost of the penalty. The choice of value for $C$ influences the classification outcome. If $C$ is too large, then the classification accuracy rate is very high in the training phase, but very low in the testing phase. If $C$ is too small, then the classification accuracy rate is unsatisfactory, making the model useless. Parameter $\gamma$ has a much greater influence on classification outcomes than $C$, because its value affects the partitioning outcome in the feature space. An excessively large value for parameter $\gamma$ results in over-fitting, while a disproportionately small value leads to under-fitting (Lin et al., 2008; Pardo and Sberveglieri, 2005).

In order to find an optimum value for these parameters, we use a metheuristic algorithm, so-called Modified Cuckoo Search (MCS) (Walton et al., 2011). MCS is recently introduced for solving optimization problems. This algorithm is based on the obligate brood parasitic behavior of some cuckoo species in combination with the Lévy flight behavior of some birds and fruit flies.

We propose a three-stage method for classification of the Diabetes diseases. At first, PCA is applied to scrap the redundant and irrelevant features. Then, MI is used to calculate the corresponding weight for each feature. Finally, MCS finds the best values for SVM parameters to classify the Diabetes diseases. Experimental results show that our proposed method called MI-MCS-SVM is able to achieve higher classification rate than what reported in previous works.

The rest of the paper is organized as follows. Section 2 presents a brief overview of Diabetes disease dataset. In section 3 we will have a look at PCA as our feature reduction method. Section 4 is devoted to SVMs. Weighted Feature Support Vector Machines (WFSVMs) based on MI is introduced in section 5. Modified Cuckoo Search is presented in section 6. Our proposed model is clarified in section 7 and finally, experimental results and conclusions are represented in section 8 and 9, respectively.

## 2 Diabetes Disease Dataset

In this study, we use the UCI Diabetes diseases dataset introduced by Black C.L. (Blake C. L., 1998). This dataset contains 768 samples, where each sample has 8 features which are eight clinical findings:

1. Number of times pregnant
2. Plasma glucose concentration a 2 h in an oral glucose tolerance test
3. Diastolic blood pressure (*mm Hg*)
4. Triceps skin fold thickness (mm)
5. 2-hour serum insulin (*mu U/ml*)
6. Body mass index (*kg/m^2*)

7. Diabetes pedigree function
8. Age (years)

These features are detailed in Table 1. All patients in this dataset are Pima Indian women at least 21 years old and living near Phoenix, Arizona, USA. The binary target variable takes the values '0' or '1'. While '1' means a positive test for Diabetes , '0' is a negative test. There are 268 cases in class '1' and 500 cases in class '0.'

| Features | Mean | Standard deviation | Min/max |
| --- | --- | --- | --- |
| 1 | 3.8 | 3.4 | 0/17 |
| 2 | 120.9 | 32.0 | 0/199 |
| 3 | 69.1 | 19.4 | 0/122 |
| 4 | 20.5 | 16.0 | 0/99 |
| 5 | 79.8 | 115.2 | 0/846 |
| 6 | 32.0 | 7.9 | 0/67.1 |
| 7 | 0.5 | 0.3 | 0.078/2.42 |
| 8 | 33.2 | 11.8 | 21/81 |

**Table 1**. Brief statistical analyze of Pima Indian Diabetes disease dataset

**3 Feature Selection**

Feature selection methods play an important role in classifying systems. They try to get rid of irrelevant and redundant features from a dataset to reduce the dimension of the dataset. It allows the complexity of data to be decreased and consequently to increase the performance of the classifying system. PCA is one of the most successful statistical techniques for dealing with this problem. It has been applied to the problem of diagnosis of Diabetes diseases by Polat and Günes (Polat and Günes, 2007). They showed that using four selected features by PCA is more promising than using all the eight features.

In our work, we apply PCA to reduce the dimension of Diabetes dataset which can be explained as following.

Assume $D$ be a $d$-dimensional dataset. The $n$ principal axes $v_1, v_2, \ldots, v_n (1 \leq n \leq\leq d)$ are orthonormal axes which their variance is maximum in the project space. Generally, $v_1, v_2, \ldots, v_n$ can be obtained by the n leading eigenvectors of the sample covariance matrix:

$$S = \left(\frac{1}{L}\right)\sum_{k=1}^{L}(x_k - m)^T(x_k - m) \qquad x_k \in D \quad (1)$$

where m is the average of samples, and L is the number of samples. Therefore,

$$Sv_k = \lambda_k v_k \qquad k \epsilon 1, \ldots, n \quad (2)$$

where $\lambda_k$ is the kth largest eigenvalue of $S$. The $n$ principal components of a given sample $x_k \epsilon D$ are given in the following

$$q = [q_1, q_2, \ldots, q_n] = V^T x_k \qquad V = [v_1, v_2, \ldots, v_n]: \quad (3)$$

where $q_1, q_2, \ldots, q_n$ are the n principal components of $x_k$.

4 Support Vector Machines

SVM classifier is a supervised learning algorithm based on statistical learning theory introduced by Vepnik (Vapnik, 1995). The main idea behind this method is to determine a hyperplane that optimally separates two classes using training dataset. Assume $\{x_i, y_i\}_{i=1}^{N}$ be a training dataset, where x is the input sample, and $y \epsilon \{+1, -1\}$ is the label of classes. Then the hyperplane is defined as $w.x + b = 0$ where x is a point lying on the hyperplane, w determines the orientation of the hyperplane, and b is the bias of the distance of hyperplane from the origin. To find the optimum separating hyperplane, $\|w\|^2$ must be minimized under the constraint $y_i(w.x_i + b) \geq 1, i = 1,2, \ldots, n$ as shown in figure 2. Therefore, it is required to solve the optimization problem given by:

$$\min \frac{1}{2}\|w\|^2$$
$$y_i(w.x_i + b) \geq 1, i = 1,2, \ldots, n \quad (4)$$

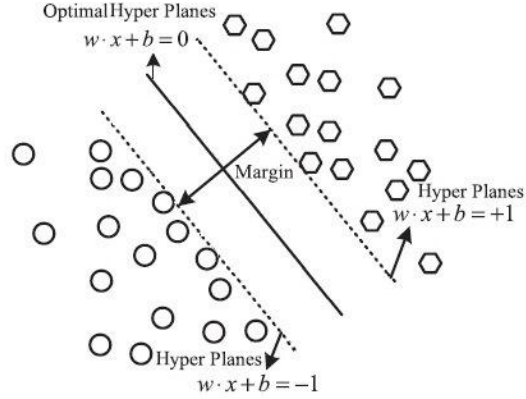

**Fig. 2.** The classification process of SVM classifier.

Now, the positive slack variables $\xi_i$ are introduced to substitute in the optimization problem and allow the method to extend for a nonlinear decision surface. The new optimization problem is given as:

$$\min_{w,\xi} \frac{1}{2}\|w\|^2 + C\sum_{i=1}^{N}\xi_i$$

$$s.t \quad y_i(w \cdot x_i + b) \geq 1 - \xi_i, \; \xi_i \geq 0, \; i = 1, 2, ..., n \quad (5)$$

where C is a penalty parameter which manages the tradeoff between margin maximization and error minimization. Thus, the classification decision function becomes:

$$f(x) = sign(\sum_{i=1}^{N} L_i y_i K(x_i, x_j) + b) \quad (6)$$

where $L_i$ are Lagrange multipliers and $K(x_i, x_j) = \varphi(x_i) \cdot \varphi(x_j)$ is a symmetric positive definite kernel function. It maps the data into a higher dimensional space through nonlinear mapping function $\varphi(x)$ for a non-linear decision system (Wei et al., 2011).

Many kernel functions have been proposed for classification problem. The most popular and widely used kernels are linear, polynomial and RBF. In our work, we use RBF kernel function to detect Diabetes diseases. Consider two samples $x_i = [x_{i1}, x_{i2}, ..., x_{id}]$ and $x_j = [x_{j1}, x_{j2}, ..., x_{jd}]$. The RBF kernel is calculated using:

$$K(x_i, x_j) = \exp\left(-\gamma \|x_i - x_j\|^2\right) \quad \gamma > 0 \quad (7)$$

where $\gamma$ is the width of Gaussian.

## 5 Weighted Feature Support Vector Machines based on Mutual Information

Traditional SVMs assume that all features of a sample contribute equally to the classification task. However, in real-world application features contribute differently in a given classification tasks. In order to contribute features differently based on their relevancy, we use the idea of weighted features using Mutual Information (MI) proposed by Xing et al. (Xing et al., 2009). Given the training set $\{x_i, y_i\}_{i=1}^{N}$ and the weighted vector $\alpha \in \mathcal{R}^d$ so that:

$$\sum_{i=1}^{d} \alpha_i = 1, \quad \alpha_i \geq 0 \quad (8)$$

Now with respect to the formula (4), the optimization problem (5) can be written as follows:

$$\min \frac{1}{2}\|w\|^2$$

s.t $y_i(w. \text{diag}(\alpha)$

where

$$\text{diag}(\alpha) = \begin{bmatrix} \alpha_1 & 0 & \cdots & 0 \\ 0 & \alpha_2 & \cdots & 0 \\ \cdots & \cdots & \cdots & \cdots \\ 0 & 0 & \cdots & \alpha_d \end{bmatrix}.$$

Substituting (8) and (9) into (5), yields the following new optimization problem:

$$\min_{w,\xi} \frac{1}{2}\|w\|^2 + C\sum_{i=1}^{N}\xi_i$$
$$s.t \ \ y_i(w\ \text{diag}(\alpha)x_i + b) \geq 1 - \xi_i\ ,\ \xi_i \geq 0,\ i = 1,2,\ldots,n$$
$$\sum_{i=1}^{d}\alpha_i = 1,\ \alpha_i \geq 0 \quad (10)$$

Finally, the classification decision function becomes:

$$f(x) = \text{sign}\left(\sum_{i=1}^{N} L_i y_i K'(x_i, x_j) + b\right) \quad (11)$$

where $K'(x_i, x_j)$ is the weighted feature RBF kernel as you can see in the following:

$$K'(x_i, x_j) = \exp\left(-\gamma\sqrt{\sum_{k=1}^{d}\alpha_k(x_{ik} - x_{jk})^2}\right) \quad (12)$$

### 5.1 Determining Feature Weights by Mutual Information

The Mutual Information (MI) between two random variables X and Y, denoted by $I(X, Y)$, is a quantity that measures the mutual dependence of these variables. Generally, when they are discrete random variables, MI is defined as:

$$I(X, Y) = \sum_{y \in Y}\sum_{x \in X} p(x, y)\log\frac{p(x, y)}{p(x).p(y)}, \quad (13)$$

where $p(x, y)$ is the conditional probability distribution function of X and Y, and $p(x)$ and $p(y)$ are the probability distribution functions of X and Y, respectively. Intuitively, MI measures the shared information between X and Y. In present work, in order to determine MI, Parzen window-based approach introduced by Kwak and Chong-Ho (Kwak and Chong-Ho, 2002) is used. Given the training dataset D with d number of features $f_1, f_2, \ldots, f_d$, the MI between each feature $f_i$ and a target class variable Y is expressed as entropy-based form as follow:

$$I(Y, f_k) = H(Y) - H(Y|f_k) \quad k = 1, 2, \ldots, d \quad (14)$$

where,

$$H(Y) = -\sum_{y} P(y)\log(P(y)), \quad (15)$$

and the probability of Y taken as y is estimated as:

$$\widehat{P}(y) = \Pr\{Y = y\} = \frac{|y|}{l} \quad (16)$$

where $|y|$ is the number of occurrence of y, and l is the total number of samples. In equation 14:

$$\widehat{H}(Y|f_k) = -\sum_{i=1}^{l}\frac{1}{l}\sum_{y=1}^{N_y} \hat{p}(y|x_i)\log\hat{p}(y|x_i), \quad (17)$$

where $N_y$ is the number of classes of $D$. Finally, the weighted vector a is determined by normal form as:

$$\alpha_k = \frac{I(Y, f_k)}{\sum_{i=1}^{d} I(Y, f_i)} \quad k = 1, 2, \ldots, d. \quad (18)$$

## 6 Modified Cuckoo Search (MCS)

MCS is a metaheuristic algorithm which is inspired by the reproduction strategy of cuckoos. At the most basic level, cuckoos lay their eggs in the nests of other host birds, which may be of different species. The host bird may discover that the eggs are not its own so they destroy the egg or abandon the nest. This has resulted in the evolution of cuckoo eggs which imitate the eggs of host birds (Yang X-S, Deb, 2010). In order to employ this technique as an optimization tool, Yang and Deb applied three idealized rules:

- Each cuckoo lays one egg, which is candidate of a set of solution co-ordinates, at a time and drops it in a random nest;
- Some of the nests containing the best eggs, or solutions, will carry over to the next generation;
- The number of nests is fixed and there is a probability that a host can discover an alien egg, say $p_a \in [0,1]$. If this happens, the host can either discard the egg or the nest and these results in building a new nest in a new location.

In MCS Walton et al. (Walton et al., 2011) try to do two modifications in order to increase the speed of the convergence of the algorithm. The first modification is regarding the step size $\alpha$ (the size of the Lévy flight). In CS, $\alpha$ equals to 1 ($\alpha=1$) (Yang and Deb., 2009) while in MCS the value of $\alpha$ decreases as the number of generations increases. An initial value of the Lévy flight step size $A = 1$ is selected and, at each generation, a new Lévy flight step is computed by $\alpha = A/G$, where $G$ is the generation number. This search is only performed on the fraction of nests to be abandoned. The next modification to original CS provides the ability of information exchange between the eggs with the hope to speed up convergence to a minimum. That is, a fraction of the eggs with the best fitness are put into a group of top eggs and for each of the top eggs, a second egg in this group is picked at random and a new egg is then generated on the line connecting these two top eggs. The distance along this line is computed using the inverse of the golden ratio $\varphi = (1+\sqrt{5})/2$. If two eggs have the same fitness, the new egg is generated at the midpoint. Meanwhile, if in this step the same egg is picked twice, local Lévy flight search is performed from the randomly picked nest with step size $\alpha = A/G^2$. The steps involved in the modified cuckoo search are shown in detail in Algorithm 1. There are two parameters, the fraction of nests to be abandoned and the fraction of nests to make up the top nests, which need to be adjusted in the MCS. Through testing on benchmark problems, it was found that setting the fraction of nests to be abandoned to 0.75 and the fraction of nests placed in the top nests group to 0.25 yielded the best results over a variety of functions.

**Algorithm 1.** Modified Cuckoo Search (MCS)

$A \leftarrow MaxLévyStepSize$
$\varphi \leftarrow GoldenRatio$
Initialize a population of *n* host nests $x_i$ ($i = 1, 2, \ldots, n$)
**for** all $x_i$ **do**
    Calculate fitness $F_i = f(x_i)$
**end for**
Generation Number $G \leftarrow 1$
**While** *NumberObjectiveEvaluations*
    < *MaxNumberEvaluations* **do**
    $G \leftarrow G + 1$
    Sort nests by order of fitness
    **for** all nests to be abandoned **do**

    Current position $x_i$
    Calculate Lévy flight step size $\alpha \leftarrow A/\sqrt{G}$
    Perform Lévy flight from $x_i$ to generate new egg $x_k$
    $x_i \leftarrow x_k$
    $F_i \leftarrow f(x_i)$
**end for**
**for** all of the top nests **do**
    Current position $x_i$
    Pick another nest from the top nests at random $x_j$
    **if** $x_i = x_j$ **then**
        Calculate Lévy flight step size $\alpha \leftarrow A/G^2$
        Perform Lévy flight from $x_i$ to generate new egg $x_k$
        $F_k = f(x_k)$
        Choose a random nest $l$ from all nests
        **if** $(F_k > F_l)$ **do**
            $x_l \leftarrow x_k$
            $F_l \leftarrow F_k$
        **end if**
    **else**
        $dx = |x_i - x_j|/\varphi$
        Move distance $dx$ from the worst nest to the best nest to find $x_k$
        $F_k = f(x_k)$
        Choose a random nest $l$ from all nests
        **if** $(F_k > F_l)$ **then**
            $x_l \leftarrow x_k$
            $F_l \leftarrow F_k$
        **end if**
    **end if**
**end for**
**end while**

## 7 Proposed Method

In this section, we describe the proposed method (MI-MSC-SVM) for the detection of diabetes diseases (see figure 3). The system works automatically in three stages
1. PCA is applied for feature reduction
2. Best feature weights are estimated using MI
3. MCS is employed for finding the optimal values for $C$ and $\gamma$.

At first, PCA method is applied to extract four features from diabetes dataset. Therefore, in feature selection stage, only large principal components were used. Afterwards, the WFSVM is applied to classify patients, the feature weights are obtained by MI and finally, the MCS algorithm is applied to find the best value for C and $\gamma$ parameters of WFSVM. We describe the training process as follows:
    1. Set up parameters of MCS and initialize the population of $n$ nests (Algorithm 1)

2. Compute the corresponding fitness function formulated by $\frac{classified}{total}$ (total denotes the number of training samples, and classified denotes the number of correct classified samples) for each particle.
3. Find the best solution using MCS

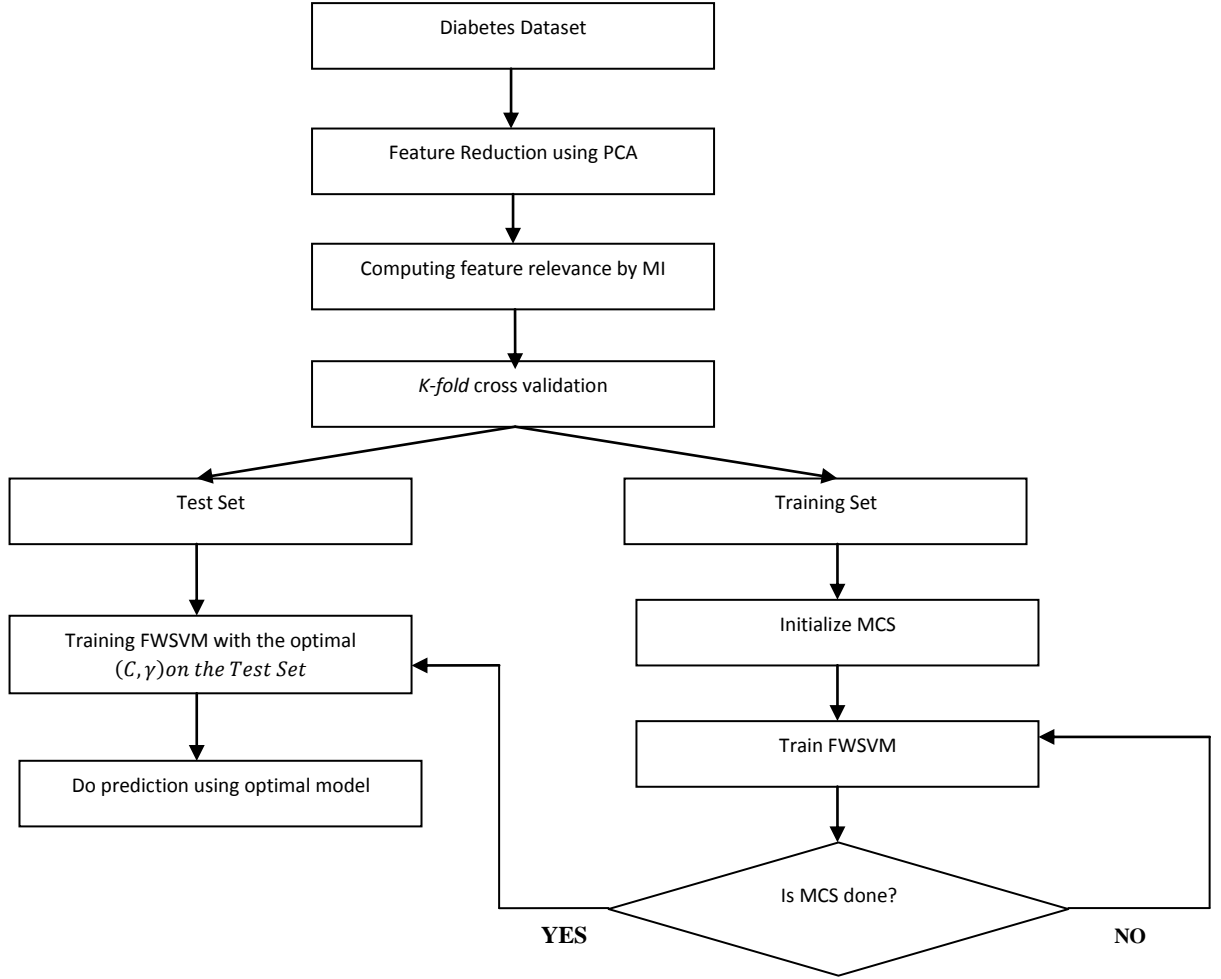

**Fig. 3**. The process of diagnosis of diabetes diseases by MI-MCS-SVM.

## 8 Experimental Results

The MI-MCS-SVM model was compared with other popular models like LS-SVM, PCA-LS-SVM, PCA-MI-LS-SVM and PCA-PSO-LS-SVM classifiers. We utilized $k$-fold cross validation to improve the holdout method. The data set was divided into $k$ subsets, and the holdout method was repeated $k$ times. Each time, one of the $k$ subsets is used as the test set and rest are put together to form a training set. Then the average error across all $k$ trials is computed (Polat and Günes, 2007). This method was applied as 10-fold cross validation in our experiments. Also, we took the related parameters of PSO in PCA-PSO-LS- SVM classifier as follows: swarm size was set to 50; the parameters $C$ and $\gamma$ were arbitrary taken from the intervals $[10^{-3}, 200]$ $and$ $[10^{-3}, 2]$, respectively. The inertia weight was 0.9, acceleration constants $C_1$ and $C_2$ were fixed to 2, and maximum number of iterations was fixed to 70. Classification results of classifiers were presented by a confusion matrix. As shown in table 2, each cell contains the raw number of exemplars classified for the corresponding combination of desired and actual system outputs.

| Output/desired | Non-diabetics | Diabetics | Methods |
| --- | --- | --- | --- |
| Non-diabetics | 44 | 6 | LS-SVM (Polat et al., 2008) |
| Diabetics | 11 | 17 | |
| Non-diabetics | 45 | 5 | PCA-LS-SVM |
| Diabetics | 9 | 19 | |
| Non-diabetics | 44 | 6 | PCA-MI-LS-SVM |
| Diabetics | 4 | 24 | |
| Non-diabetics | 45 | 5 | PCA-PSO-LS-SVM |
| Diabetics | 4 | 24 | |
| Non-diabetics | 48 | 2 | MI-MCS-SVM |
| Diabetics | 3 | 25 | |

**Table 2**. Confusion matrix

In this way one can observe the frequency by which a patient is misclassified. In addition, Table 3 shows the classification accuracies of MI-MCS-SVM. The proposed method achieved the highest classification accuracy of 93.58% among classifiers on the test set. The test performance of the classifiers is also determined by the computation of specificity and sensitivity which are defined as:

*Specificity*: number of true negative decisions / number of real negative cases
*Sensitivity*: number of true positive decisions / number of real positive cases.

A true positive decision occurs when the positive prediction of the network coincided with a positive prediction of the physician. A true negative decision occurs when both the network and the physician suggested the absence of a positive prediction.

| Methods | Sensitivity (%) | Specificity (%) | Classification accuracy |
| --- | --- | --- | --- |
| LS-SVM (Polat et al., | 73.91 | 80 | 78.21 |
| PCA-LS-SVM | 79.16 | 83.33 | 82.05 |
| PCA-MI-LS-SVM | 80 | 91.66 | 87.17 |
| PCA-PSO-LS-SVM | 82.75 | 91.83 | 88.46 |
| IM-MCS-FWSVM | 92.59 | 94.11 | 93.58 |

**Table 3.** The values of the statistical parameters of the classifiers

According to Table 4, it can be seen that using FWSVM classifier with MI and MCS, one can obtain much better classification accuracy than previously introduced methods. So, we conclude that our method is able to gain a high rate of accuracy in diagnosis of Diabetes disease. This method can be coupled with software to assist the physicians for making the final decision with more ease and confidence.

| Author | Method | Classification accuracy |
| --- | --- | --- |
| Ster et al. | QDA | 59.5 |
| Zarndt | C4.5 rules | 67 |
| Yildirim et al. | RBF | 68.23 |
| Bennet et al. | C4.5 (5xCV) | 72 |
| Zarndt | Bayes | 72.2 |
| Statlog | Kohonen | 72.8 |
| Ster et al. | ASR | 74.3 |
| Shang et al. | DB-CART | 74.4 |
| Friedman | Naïve Bayes | 74.5 |
| Zarndt | CART DT | 74.7 |
| Statlog | BP | 75.2 |
| Ster et al. | SNB | 75.4 |
| Ster et al. | NB | 75.5 |
| Grudzinski | kNN | 75.5 |
| Zarndt | MML | 75.5 |
| Statlog | RBF | 75.7 |
| Ster et al. | LVQ | 75.8 |
| Friedman | Semi-Naïve Bayes | 76 |
| Ster et al. | MLP + BP | 76.4 |
| Ster et al. | FDA | 76.5 |
| Ster et al. | ASI | 76.6 |
| Statlog | SMART | 76.8 |
| Bennet et al. | GTO DT (5xCV) | 76.8 |
| Yildirim et al. | BFGS quasi Newton | 77.08 |
| Yildirim et al. | LM | 77.08 |
| Statlog et al. | LDA | 77.5 |
| Yildirim et al. | GD | 77.6 |
| Bennet et al. | SVM (5xCV) | 77.6 |
| Polat et al. | GDA-LS-SVM | 79.16 |
| Yildirim et al. | GRNN | 80.21 |
| Calisir et al. | LDA-MWSVM | 89.74 |
| Giveki et al. (this | MI-MCS-SVM | 93.58 |

**Table 4**. Classification accuracy of our method with other methods from litrerature

**9 Conclusions**

This work presents a novel automatic model to diagnose Diabetes disease based on Feature Weighted Support Vector Machines and Modified Cuckoo Search. In order to discard the irrelevant features, Principal Component Analysis was used. Then, Mutual Information was applied to the selected features to weight them based on their relevancy to the given classification task. Results show that it improves the accuracy of the model. Moreover, Modified Cuckoo Search is used which not only speeds up the convergence of the algorithm but also allows to find the optimal values for parameters of SVM. Results demonstrate that the proposed model is faster and significantly more reliable than the models proposed in previous works. This method can be coupled with medical softwares to assist physicians to make more accurate decisions about Diabetes disease.